# A Biomedical Pipeline to Detect Clinical and Non-Clinical Named Entities

Shaina Raza
Ontario Agency for Health Protection and Promotion, Dalla Lana School of Public Health, University of Toronto
Toronto, ON, Canada
Shaina.raza@oahpp.ca

Brian Schwartz
Ontario Agency for Health Protection and Promotion, Dalla Lana School of Public Health, University of Toronto
Toronto, ON, Canada
Brian.schwartz@oahpp.ca

## ABSTRACT

There are a few challenges related to the task of biomedical named entity recognition, which are: the existing methods consider a fewer number of biomedical entities (e.g., disease, symptom, proteins, genes); and these methods do not consider the social determinants of health (age, gender, employment, race), which are the non-medical factors related to patients' health. We propose a machine learning pipeline that improves on previous efforts in the following ways: first, it recognizes many biomedical entity types other than the standard ones; second, it considers non-clinical factors related to patient's health. This pipeline also consists of stages, such as pre-processing, tokenization, mapping embedding lookup and named entity recognition task to extract biomedical named entities from the free texts. We present a new dataset that we prepare by curating the COVID-19 case reports. The proposed approach outperforms the baseline methods on five benchmark datasets with macro-and micro-average F1 scores around 90, as well as our dataset with a macro-and micro-average F1 score of 95.25 and 93.18 respectively

## 1. INTRODUCTION

Recent years have seen a dramatic increase in the number of biomedical documents (research papers, case reports, electronic health records, and clinical notes). There are approximately 28 million articles in the MEDLINE database to-date [1]. Hundreds of articles have been published in the last two years as a result of COVID-19 research [2]–[4]. As demand for biomedical knowledge grows, large-scale data management is essential. Unstructured (free) texts are extremely difficult for researchers to manage and infer information from [5]. These challenges include extracting key information from scientific texts, categorizing articles, and enabling efficient content discovery. Text mining is a subtask of Natural Language Processing (NLP) that converts free texts into a format suitable for data analysis and to build machine learning (ML) models [6].

Named Entity Recognition (NER), an NLP task, is the process of identifying and classifying key information (such as a person, organization, or event) in a text [7], and it is a key technique in text mining. In the biomedical domain, the NER task can be used to identify biomedicine entities such as genes, diseases, species, chemicals, etc. [8]. Biomedical NER research is currently focused on a small number of named entities (disease, genes, proteins, etc.) [9]. Nonetheless, there are several biomedical entities to be considered, including disease, diagnosis, medical concepts, risks, vital signs and so on, that need to be identified from texts, which is the driving force behind this research.

There are also some non-medical factors, such as Social Determinants of Health (SDoH) [10], which refer to the conditions in the environments where people are born, live, work, and grow that influence the health of populations. These factors usually have long-term effects on individual health outcomes and lead to health disparities [11], if not considered. In this study, we also place a focus on SDoHs, which is a relatively under-researched area in both the field of healthcare and Artificial Intelligence (AI).

In this work, we propose a trainable ML pipeline that extracts the biomedical named entities from the free texts. We refer to both clinical and non-clinical entities as "biomedical entities" in this paper. The specific contributions of this research are as:

- We propose and develop a Bio-Medical NER Pipeline (BMNP), to identify biomedical entities from the scientific texts. This pipeline consists of multiple data transformations and ML models (tokenizer, pre-processor, named entity recognizer). Each individual stage of the pipeline can be treated as a separate, manageable component that can be independently deployed, optimized, configured, and automated.

- We construct a new dataset by curating and scientifically analyzing a significant number of COVID-19 case reports. A case report details the symptoms, diagnosis, treatment, and follow-up of a patient. A portion of this dataset is also annotated with biomedical entities by the experts to provide a gold-standard dataset used to train and evaluate the named entity model.

- In order to ensure that we are in compliance with the Personal Health Information Protection Act (PHIPA) [12], we de-identify the patients' identifiable information (name, address, etc.) in the data after identifying the named entities. In this work, we do not have access to the personal information of actual patients; as a result, we are using the fake identifiers to construct and test this module.

We compare the effectiveness of our NER approach to state-of-the-art methods on publicly available benchmark datasets and our COVID-19 case reports dataset. The experimental results demonstrate that the knowledge transfer through this biomedical pipeline results in a substantial increase in biomedical information compared to current biomedical NER methods.

## 2. RELATED WORK

Traditional NER methods only consider specific entities (e.g., persons, organizations, locations, etc.) [7]. Biomedical NER [13] is the task of identifying entities in the biomedical domain, such as chemical compounds, genes, proteins, viruses, disorders, drugs, adverse effects, diseases, DNAs and RNAs. In the state-of-the-art of biomedical NER, most of the research [9], [14] focuses on general approaches to named entities that are not specific to the biomedical field. On the other hand, there are some works [15], [16] that focus solely on biomedical and chemical NER, however, they do not cover many clinical entities, such as diseases, symptoms, clinical procedures and such. SDoHs [17] also a major impact on people's health, and well-being and are related to health outcomes, which is rather an underexplored research area in biomedicine research.

According to a 2016 survey, about 95% of U.S. hospitals use EHRs [18]. Case reports [19] also contain patients' data that can be used as a substitute for EHRs [20] and are distributed for free for research purposes. The term "de-identification" refers to the process of removing or replacing personal identifiers in such a way that re-establishing a link between this information should not be possible. Some studies employ de-identification as a sub-task of the biomedical NER task [21], where patient personal entities are recognized first and are then de-identified.

The Conditional Random Field (CRF) models [22], and Structured Support Vector (SVM) [23] are commonly used models for NER, biomedical NER and de-identification tasks. Deep learning models based on Recurrent Neural Networks (RNN) and Convolutional Neural Network (CNN) are also used for biomedical named entities and deidentification purpose [24]. In recent times, BioBERT[16], SciBERT[25] and related Transformer-based models are also used to identify named entities from biomedical texts.

Deep neural language models, such as Transformer based models [26], have recently evolved to a successful method for representing text. In particular, Bidirectional Encoder Representations from Transformers (BERT) outperformed previous state-of-the-art methods by a large margin on various NLP tasks [27]. BERT has been adopted for a number of scientific, including the biomedicine field. For example, BioBERT [16] is a well-known adaptation of BERT for the biomedical domain, pre-trained on PubMed abstracts and PMC full-text articles.

COVID-Twitter-BERT (CT-BERT) is another Transformer-based model, pretrained on a large corpus of Twitter messages on the topic of COVID-19. BioRedditBERT [28] is also BERT model initialized from BioBERT weights and further pre-trained on health-related Reddit posts.

In this work, we also use deep learning-based methods to build a pipeline for the biomedical NER and de-identification tasks. Our pipeline contains both the use of Transformer-based models for embeddings and a deep neural network-based BiLSTM and CRF architecture. Different from previous works, we extend the standard biomedical NER to identify many named entities. We also include the clinical BERT embeddings [29] in this work.

## 3. METHODOLOGY

We develop a ML pipeline that we name Bio-Medical Named entity recognition Pipeline (BMNP), shown in Figure 1.

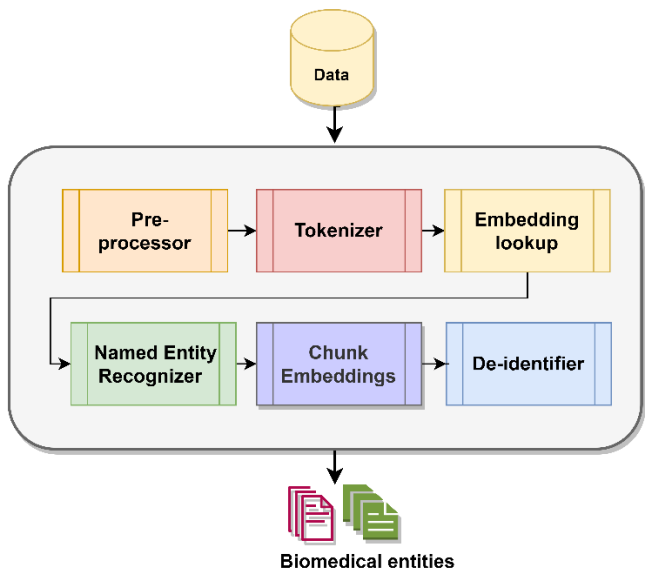

**Figure 1:** Bio-Medical Named entity recognition Pipeline

We build this pipeline following Spark pipeline [30] format, which takes raw data, pre-processes it, and applies algorithms to recognize and classify biomedical entities into predefined classes. Next, we discuss each component of this pipeline in detail. The data (input) and biomedical entities (output) are explained in Section 4.

The BMNP pipeline consists of following components:

**Pre-processor:** The pre-processor pre-processes the input data and detects the sentence boundaries in each document. Then, it transforms the data into a format that is readable by the next stage in the pipeline. The output from this stage is a set of pre-processed documents.

**Tokenizer:** The pre-processed data from the previous stage (i.e., pre-processing) goes to the tokenization stage, that is handled by the tokenizer. Tokenization is a process of breaking the input text into smaller chunks (words/ sentences) called tokens [31]. The output from this stage is a transformed data, containing tokens (words) corresponding to each document.

**Embedding lookup:** The tokenized data from the tokenizer goes into the embedding lookup stage, which is handled by the embedding lookup component. We have used the BERT embeddings [29] pre-trained on PubMed corpora and MEDLINE. This embedding lookup node maps tokens to vectors. This component can also download other pre-trained embeddings (such as Glove, BERT, BioBERT, etc.,). The output from this stage are the embeddings corresponding to each word in the document.

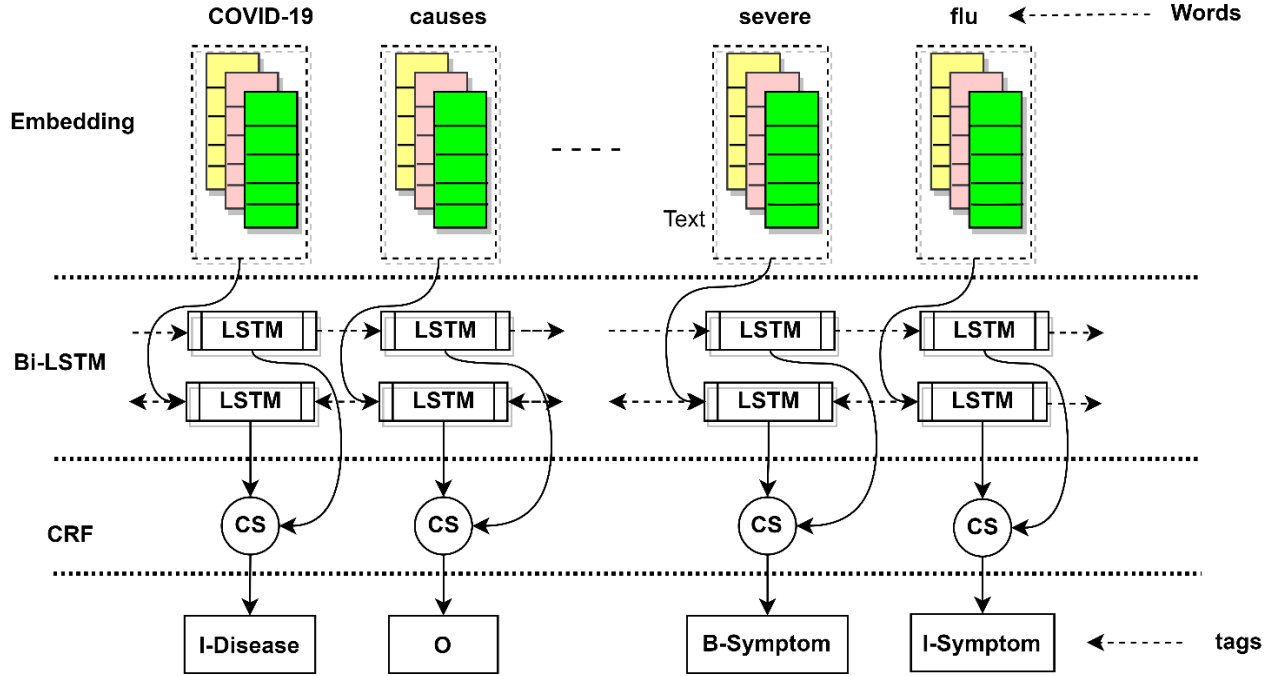

**Figure 2**: BiLSTM-CRF algorithm

**Named Entity Recognizer:** Given an ordered set of *N* character sequences and an ordered set of annotations, the task is to create a predictor, where the output is a set of biomedical entities similar to the actual annotations. This component can identify the biomedical entities in the documents. We adapt the Bi-directional Long short-term memory (BiLSTM) - Conditional Random Field (CRF) [32] model for the biomedical NER task. We refer to this model as BiLSTM-CRF model. We show the working of BiLSTM- CRF in Figure 2 and explain its work next.

This component consists of four layers, which are (i) embedding layer; (ii) BiLSTM layer; (iii) CRF layer and (iv) output layer. As shown in Figure 2, the BiLSTM-CNN-CRF algorithm takes a sequence of words as $s = [w_1, w_2, \ldots, w_N]$ as input, where $w_i$ refers to the one-hot representation of the $i^{th}$ word in the sequence. This input goes to the first layer, which is the embedding layer.

*Embedding Layer:* The embedding layer consists of the two representations: (i) word embedding features and (ii) character features.

We use the pre-trained embeddings that are loaded and provided by the embedding lookup component to get the word embeddings. The pre-trained models provide embeddings for both medical and non-medical terms.

We use Convolution Neural Network (CNN) for producing character embedding. Character-level word embedding is useful to address the issue of out-of-vocabulary words from the word embedding. We use CNN network to capture local information within given words in a biomedical context. Each position in the sequence has sliding windows, and CNN performs a transformation for each sliding window. The contextual representation $c_i$ of the $i_{th}$ character is learned by using the CNN filter, as shown in Equation (1):

$$c_i = f(w^T \oplus x_{[i \pm \frac{K-1}{2}]}) \quad (1)$$

where $x_i$ is the dense vector representation of word $w_i$, $x_{[i \pm \frac{K-1}{2}]}$ represents the concatenation of embeddings of characters from $[i - \frac{K-1}{2}]$ to $[i + \frac{K-1}{2}]$. We use the Rectified Linear Activation Unit (ReLU) as an activation function *f*. The contextual representation $c_i$ is the concatenation of the outputs of all filters at this position. The output of CNN layer is $c = [c_1, c_2, \ldots, c_N]$, where $c_i \in R^M$, *M* refers to the number of filters in CNN layer.

*Bi-LSTM layer:* The second layer in the model is the Bi-LSTM network. Once, we concatenate the feature representations in the embedding layer, we make use of the BiLSTM layer to learn sequential structure of words using all previous contexts (in both directions). The hidden representation $h_i$ is a concatenation of $\vec{h}_i$ and $\overleftarrow{h}_i$ as $[\vec{h}_i, \overleftarrow{h}_i]$. The output of Bi-LSTM layer is $h = [h_1, h_2, \ldots, h_N]$, where $h_i = R^{2S}$, where *S* refers to the dimension of hidden states in LSTM.

*CRF layer*: The third layer on the top of the Bi-LSTM network is the CRF layer [33]. The input to the CRF layer is $h = [h_1, h_2, \ldots, h_N]$ generated by the Bi-LSTM layer, where *h* refers to the sequence of hidden states. CRF is a conditional probability distribution model mostly used in sequence labelling tasks to generate new tags based on previously labelled tags [33].

The output of the CRF layer is $y = [y_1, y_2, \ldots, y_N]$, where *y* refers to the sequence of labels. In this work, the biomedical entities are the labels. A tanh layer on top of the BiLSTM layer is added to predict the confidence scores (CS) for every word with each of the possible labels as the output score of the network, as shown in Equation (2):

$$CS_i = tanh(W_c h_i + b_c) \quad (2)$$

where $W_c$ and $b_c$ are trainable parameters. In training, we use the negative loglikelihood function over all training samples to calculate the loss function $\mathcal{L}$, which is shown in Equation (3) as:

$$\mathcal{L}_{BMNP} = -\sum_{s \in \mathbb{S}} log(p(y_s | h_s; \theta)) \quad (3)$$

Where $\mathbb{S}$ refers to the set of sentences in training data, $p$ denotes the probability and $\theta$ refers to the parameters during training.

*Output layer:* This layer converts the IOB representation of named entities to a user-friendly representation, by associating the tokens of recognized entities and their labels. Each output from this stage is a 'chunk' that is a tagged portion of sentence into named entities.

**Chunk Embeddings:** This component combines the embedding of a chunk with the embedding of the surrounding. We load the BERT-based clinical embeddings [29] for both the chunk and the sentence, through embedding lookup component. For sentence embeddings, we use the weights of BERT clinical embeddings to get the sentence-level representations using Sentence Transformers[1]. For each input chunk annotation, the chunk embedding component identifies the corresponding sentence, computes the BERT sentence embedding for both the chunk and the sentence, and then takes the average. The output embeddings are useful when a context-sensitive representation of a text chunk is needed.

**De-identifier:** The input to this component are the chunks associated with the tagged entities and output is the text that is de-identified. In this component, we employ the data obfuscation technique, which is a process that conceals the meaning of data [34]. For example, to replace identified names with different fake names or to mask some data value <15-06-2022> with <DATE>. This component provides PHIPA compliance when dealing with text documents containing any protected health information. We use the pre-trained de-identification model[2] from John Snow Labs inside the pipeline to de-identify patients' records.

## 4. EXPERIMENTS
### 4.1 Data

**Our COVID-19 case reports dataset:** We have collected the clinical case reports from different journals (Lancet, BMJ, AMJ, Clinical Medicine and other related journals) that are standardized according to the CAseREports (CARE) guidelines [19]. The inclusion criteria are given below:

- We include only the PubMed Central (PMC) case reports that are in English Language between March 20, 2021 and March 20, 2022 for data collection.
- We exclude many early-pandemic case reports, as the disease symptoms, diagnosis, drugs, and vaccination information were unclear at that time.

We scrap the PDFs of these case reports and use Apache Tikka toolkit to extract metadata (authors' names, DOI, journal name, case report title) and full texts from these documents. After completing these steps, we found around 4500 case reports.

**Gold-standard dataset:** Gold-standard dataset [35] means a corpus of text or a set of documents that are manually annotated with the labels. We use the Label Studio annotation tool[3] to annotate around 500 case reports and prepare it in the CONLL [36] format to construct a gold-standard dataset; which according to research [37], is good number to begin training an NLP model. We train the BiSLTM-CNN-CRF algorithm inside the pipeline using this gold-standard dataset to initiate the biomedical NER task.

**Benchmark datasets:** We use the following benchmark datasets to train the NER component: JNLPBA [38], NCBI-Disease[39], BC5CDR[40], BC2GM[41], miRNA [42].

These datasets are readily available in CoNLL-2003 format here[4], which is a kind of prototypical standard for building algorithms that recognize named entities in the texts [36]. We performed additional processing to convert the datasets into IOB (Inside-Outside-Before) scheme, which is used for sequence labelling. It is also facility provided by our pipeline. We summarize the details of datasets used in this study in Table 1. The benchmark datasets are mostly based on PubMed abstracts.

**Table 1.** Datasets used in this work

| Dataset | Entity type | Annotat-ions | Data size |
|---|---|---|---|
| JNLPBA | Gene/ Protein | 35,336 | 2,404 abstracts |
| NCBI | Disease | 6,881 | 793 abstracts |
| BC5CDR | Chemicals | 15,935 | 1,500 articles |
| BC2GM | Gene/ Protein | 24,583 | 20,000 sentences |
| miRNA | Disease, Organisms, Genes/proteins | ~14000 | ~7000 abstracts |
| Case reports | Clinical and non-clinical | ~25,000 | 4500 case reports |

### 4.2 Biomedical Named Entities

We get biomedical named entities as the output of the BMNP. We include a number of clinical and non-clinical entities that we finalized after reviewing the relevant literature [43], [44]. These entities are listed as

**Clinical entities:** Admission (patient admission status), oncology (tumor/cancer), blood pressure, respiration (shortness of breath), dosage (medicine), vital signs, symptoms, kidney disease, temperature (body), diabetes, vaccine, time of symptom (days, weeks), obesity, pregnancy, BMI, height (of patient), heart disease, pulse, hypertension, drug name, drug ingredient, hyperlipidemia, cerebrovascular disease, disease syndrome disorder, treatment, clinical department, weight (of patient), admission/ discharge (from hospital), modifier (modifies current state), external body part, test, strength, route, test result, drug.

**Non-clinical entities:** Name (of patient), location, date, relative date, duration, relationship status, social status, family history (family members, alone/ with family/ homeless), employment status, race/ethnicity, gender, sexual orientation, diet (food type, nutrients, minerals), alcohol, smoking.

---

[1] https://www.sbert.net/

[2] deidentify_rb_en_2.0.2_2.4_1559672122511.zip

[3] https://labelstud.io/

[4] CONLL format Datasets

### 4.3 Experimental Settings

We use PyTorch for the implementation of models. We configure this pipeline using the Spark NLP configurations [30], which enables us to scale up in clusters while adhering to distributed data processing principles. It also supports in-memory distributed data processing, which results in faster training and inference.

We run our experiments on Google Colab Pro and used Apache Spark NLP in local mode (no cluster) to integrate the components of the ML pipeline. We specify the following hyperparameters as shown in Table 2.

We use Grid search to get the optimal values for the hyperparameters and early stopping to overcome possible overfitting. We have divided all datasets into training, validation, and test sets, with a 70:15:15 ratio. We used the Stratified 5-Folds cross-validation (CV) strategy for train/test split if original datasets do not have an official train/test split.

**Table 2.** Hyperparameters used

| Hyperparameter | Optimal value (values used) |
| --- | --- |
| Learning rate | 1.E-03 (1.E-02, 1.E-03, 1.E-05, 2.E-05, 5.E-05, 3.E-04) |
| Batch size | 64 (8, 16, 32, 64, 128) |
| Epochs | 30 ({2, 3, …, 30}) |
| LSTM state size | 200 (200, 250) |
| Dropout rate | 0.5 ({0.3, 0.35, …., 0.7}) |
| Optimizer | Adam |
| CNN filters | 2 (2,3,4,5) |
| Hidden Size | 768 |
| Embedding Size | 128 |
| Max Seq Length | 512 |
| Warmup Steps | 3000 |

**Evaluation metrics**: Following the standard practice [45] to evaluate NER tasks, we use the following metrics:

- Micro-average F1 measures F1-score of aggregated contributions of all classes.
- Macro-average F1 computes the arithmetic mean of all the per-class F1 scores.

**Baseline methods:** We test the performance of our BMNP approach against the following methods:

- SciBERT[25]: we use the implementation of allenai/scibert-base pre-trained on biomedical data with ~785k vocabulary.
- BioBert[16]: BioBERT is a pre-trained language model for biomedical text mining. We use the BioBERT-base-cased with following versions:
- BioBert v1.0 pre-trained on 200k PubMed articles.
- BioBERT v1.1 pre-trained on 1M PubMed articles.
- BioBERT v1.2 pre-trained on 1M PubMed articles in the same way as BioBERT v1.1 but includes a language modelling (LM) head.
- CT-BERT [46]: it is a BERT-large-uncased model, pretrained on Twitter messages on the topic of COVID-19.
- BiLSTM-CRF [47]: we use a standard BiLSTM-CRF architecture that relies on contextual string embeddings.

All the baselines are trained on the datasets discussed above. Each baseline is tuned to its optimal hyperparameter setting and the best results for each baseline are reported.

## 5. RESULTS
### 5.1 Comparison with Baselines

We report the results of all methods on all datasets using macro average F1 (macro) and micro average F1 (micro) scores in Table 3. These scores show the percentage values. Bold means highest performance.

Overall, these results in Table 3 show that our BMNP approach achieves the best performance on five public biomedical benchmarks (NCBI Disease, BC5CDR, JNLPDP, BC2GM, MiRNA) as well as on our case-reports designed specifically for biomedical named entities. This shows the usefulness of our methodology across different types of datasets. We see the biggest performance boost when our pipeline is tested on our case reports dataset that is annotated with many biomedical named entities.

This superiority of our approach is attributed to two important things: (1) the embedding lookup component that can load the domain-specific pre-trained language model (we use clinical BERT embeddings) to get the relevant embeddings. (2) Our approach stacks together various ML components (Figure 1), where each node prior to NER component contributes to identifying the biomedical entities and the chunk embedding component help us to assign chunks by looking into the context of the sentences.

Our approach achieves the best micro of 93.14 on our dataset (around 52 entities), 91.30 on miRNA (disease, organism, gene. Proteins), 91.12 on NCBI Disease (disease entity), 89.12 on BC5CDR (chemicals), 89.15 on BC2GM (gene/proteins) and 90.13 on JNLPDP (gene/proteins) dataset. We also observe the best macro of 95.25 on our dataset, 92.93 on miRNA, 92.14 on NCBI, 90.23 on BC2GM, 89.14 on BC5CDR, and 89.01 on JNLPDP dataset.

The BioBERT model also shows a competitive performance in these results. We find that BioBERT achieves better performance on disease entities (NCBI), followed by chemical (BC5CDR) and then gene/proteins entities (miRNA, BC2GM and JNLPDP). It performed very well on our case reports dataset. BioBERT is a state-of-the-art method that has been fine-tuned in many follow-up work in biomedicine. Among its variants, we see the overall better performance of BioBERT v1.2 than its other predecessors, except for a few places, where BioBERT v1.1 marginally outperforms BioBERT v1.2. This was expected since BioBERT v1.2 includes a language model head, during the training, which can be useful for the classification problems. In this work, we use the clinical BERT embedding pre-trained on PubMed and Medline data, which shows at least 1-3% better results than BioBERT pre-trained on PubMed abstracts.

We also observe the better performance of BiLSTM-CRF model in identifying many biomedical entities in these experiments. The BiLSTM-CRF, though not as deeper as a BERT model, performs better than SciBERT and CT-BERT.

|  | Datasets | | | | | |
|---|---|---|---|---|---|---|
| Model | NCBI | BC5CDR | JNL PDP | BC2GM | miRNA | Case reports |
| micro F1 average | | | | | | |
| CT-BERT | 62.67 | 62.91 | 60.27 | 62.82 | 67.23 | 68.16 |
| SciBert | 81.15 | 80.72 | 77.13 | 76.78 | 74.34 | 76.23 |
| BiLSTM-CRF | 83.32 | 83.92 | 79.23 | 78.04 | 78.35 | 81.23 |
| BioBert v1.0 | 86.01 | 84.56 | 78.68 | 85.28 | 84.24 | 85.87 |
| BioBERT v1.1 | 88.52 | 87.15 | 79.39 | 86.16 | 85.24 | 86.27 |
| BioBERT v1.2 | 89.12 | 87.81 | 83.34 | 86.45 | 85.98 | 86.88 |
| BMNP (ours) | **91.12** | **89.12** | **90.13** | **89.15** | **91.30** | **93.14** |
| macro F1 average | | | | | | |
| CT-BERT | 63.14 | 63.24 | 61.15 | 63.23 | 63.10 | 68.72 |
| SciBert | 82.13 | 79.88 | 80.65 | 80.13 | 78.00 | 78.29 |
| BiLSTM-CRF | 84.12 | 84.02 | 83.56 | 79.32 | 77.10 | 78.10 |
| BioBert v1.0 | 79.10 | 78.90 | 79.00 | 78.13 | 70.21 | 78.18 |
| BioBERT v1.1 | 85.89 | 87.10 | 87.18 | 85.45 | 83.88 | 87.78 |
| BioBERT v1.2 | 86.78 | 87.89 | 86.07 | 85.15 | 84.80 | 86.98 |
| BMNP (ours) | **92.14** | **89.14** | **89.01** | **90.23** | **92.93** | **95.25** |

**Table 3.** Performance of methods on different datasets using micro and macro

SciBERT is initially trained on scientific data (not clinical), its performance is somehow compromised on biomedical entities. In the same way, CT-BERT is pre-trained on Twitter data, so the meanings of entities are different from pure biomedical entity types.

Our modified BiLSTM-CRF with CNN layer performs better than the vanilla BiLSTM-CRF baseline, probably because, we are using pre-trained biomedical embeddings in the embedding lookup. We also use the chunk embeddings (consisting of BERT sentence level embeddings) for chunk annotations which consider the context of the sentence the chunk (e.g., a tagged phrase or word) in the text. We are also using a deeper neural network - with more layers than standard BiLSTM-CRF.

Although we fine-tune each baseline method to its optimal hyperparameter settings, we anticipate that the relatively low scores of these baselines on our case reports dataset can be attributed to the following: (i) absence of a training dataset for training new biomedical, and (ii) different training/test set splits used in previous works that were unavailable.

Compared to previous research, our method is capable of identifying a wide range of clinical and non-clinical entity types. We extract various entries pertaining to medical risk factors (hypertension, kidney, diabetes, etc.), patients' personal information (age, gender, geography), SDoH (diets, race, income), and other clinical entity types, including underlying disease, tissue, and organ systems. The majority of biomedical projects focus on chemicals, proteins, and genes; however, our pipeline is highly adaptable and capable of identifying numerous identities. This is evidenced by the fact that our pipeline performed best when trained on other datasets. When we trained our pipeline on the dataset of case reports, it again demonstrated the best performance.

## 5.2 Effectiveness of BMNP on Case Reports

We give a random snippet from a COVID-19 case report to our pipeline and show the confidence scores (Equation 2) for the predicted biomedical entities. The expectation over here is that for a given confidence score, the model should predict with a higher confidence score. The results are shown in Table 4.

**Table 4:** Confidence scores of predicted biomedical entities (sen. for sentence, beg. for beginning conf. for confidence score)

| Sen. | Beg. | End | Chunks | Biomed Entity | Conf. |
|---|---|---|---|---|---|
| 0 | 2 | 12 | 36 years old | Age | 1.00 |
| 0 | 14 | 18 | man | Gender | 1.00 |
| 0 | 32 | 43 | fever | Symptom | 0.98 |
| 0 | 52 | 65 | first hospital | Clinical Depart. | 0.51 |
| 0 | 156 | 160 | cough | Symptom | 0.99 |
| 0 | 178 | 196 | breath shortness | Symptom | 0.39 |
| 0 | 233 | 244 | 5 days | Relative Date | 0.42 |
| 1 | 247 | 249 | he | Gender | 1.00 |
| 1 | 261 | 264 | severe | Modifier | 0.90 |
| 1 | 266 | 273 | diarrhea | Symptom | 1.00 |
| 1 | 280 | 289 | stools | Symptom | 0.85 |
| 1 | 292 | 303 | 2 days after | Relative Date | 0.68 |

As seen in Table 4, our pipeline can predict many biomedical entities from the input text. Due to brevity reason, we only present a snippet of a case report, so these predicted entities do not encompass all the biomedical entities that we have defined. However, this pipeline can run over many documents and can extract the named entities.

We also show five most common clinical as well as non-clinical entities predicted using our approach from 50 case reports. Due to space limitation, we only show a few top entities in Table 5.

We observe in Table 5 that we get some valuable information regarding the most frequent clinical entities, such as diseases, symptoms, and drugs. According to this table, the most common symptom is cough and cold while the most common diseases or clinical entities mentioned are coronavirus and cardiogenic. We also observe a few non-clinical entities and find some non-clinical factors that can be analyzed to study the social impacts on population health.

**Table 5:** Most used clinical and non-clinical entities in 50 case reports.

| Clinical entities | | | |
|---|---|---|---|
| **Admission** | **Drug** | **Disease** | **Symptom** |
| hospitalized | antibiotics | coronavirus | cough |
| admitted | dobutamine | cardiogenic | cold |
| admittance | oseltamivir | Covid-19 | congestion |
| post-admission | fluids | pneumonia | abdominal |
| discharged | saline | hepatitis | hemorrhage |
| Non-clinical entities | | | |
| **Race/Ethnicity** | **Relationship** | **Age** | **Gender** |
| caucasian | single | 18-year-old | female |
| asian | married | 59-year-old | he |
| black | divorced | teenager | his |
| white | partner | older | man |

We choose a published case report [48] and show a sample prediction on case 1 on Figure 3. As shown in Figure 3, we identify many clinical as well non-clinical named entities from the given text. This is particularly useful for the medical experts to quickly go through the case histories of the patients. We can also run this pipeline on many case reports to infer some statistics on population health.

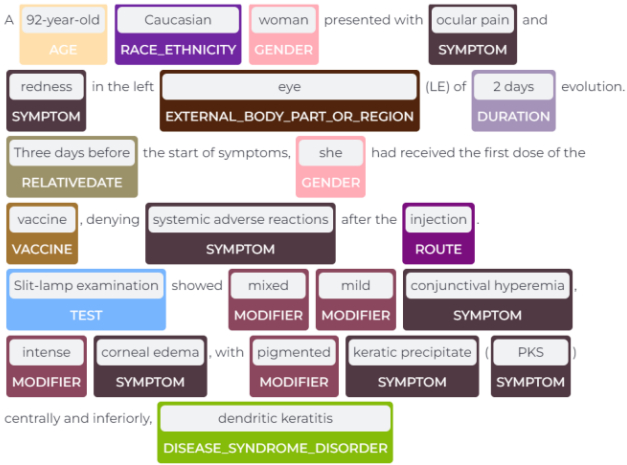

**Figure 3**: Biomedical entities based on a case report [48]

We also show the de-identification of personal information in Figure 4.

**Figure 4**: De-identification task

We have specified that following named entities that should be de-identified: age, contact, date, patient ID, location, name, profession, city, country, doctor, hospital, medical record, organization, patient, phone, profession, street, username, zip, account and license, although they can be more. We also show the entity tagging in IOB format on a clinical text in Table 6.

**Table 6:** IOB-Tagging example with precision (prec.), recall and F1-score

| Entity tagging | prec. | recall | F1-sore |
|---|---|---|---|
| B-Cerebrovascular | 0.92 | 0.92 | 0.92 |
| I-Vaccine | 0.91 | 0.74 | 0.82 |
| I-N_Patients | 0.50 | 0.29 | 0.36 |
| B-Heart_Disease | 0.80 | 0.74 | 0.77 |
| B-Obesity | 1.00 | 1.00 | 1.00 |
| B-Dosage | 0.64 | 0.54 | 0.58 |
| B-Hypertension | 1.00 | 1.00 | 1.00 |
| I-Stage | 0.00 | 0.00 | 0.00 |
| I-Cell_Type | 0.73 | 0.83 | 0.77 |
| B-Admission_Discharge | 0.99 | 0.96 | 0.97 |
| B-Date | 0.90 | 0.91 | 0.90 |
| I-Admission_Discharge | 0.00 | 0.00 | 0.00 |
| I-Drug_Ingredient | 0.59 | 0.65 | 0.62 |
| I-Virus | 0.68 | 0.70 | 0.69 |
| B-BMI | 0.82 | 0.82 | 0.82 |
| B-Drug_Ingredient | 0.80 | 0.80 | 0.80 |
| B-Severity | 0.66 | 0.73 | 0.69 |
| I-Treatment | 0.83 | 0.52 | 0.64 |
| I-Pulse | 0.50 | 1.00 | 0.67 |
| I-Respiration | 0.33 | 1.00 | 0.50 |
| B-Death_Entity | 0.82 | 0.70 | 0.76 |
| B-Race_Ethnicity | 1.00 | 1.00 | 1.00 |
| I-Hypertension | 1.00 | 1.00 | 1.00 |
| I-Age | 1.00 | 0.83 | 0.91 |
| B-Employment | 0.73 | 0.78 | 0.75 |
| B-Smoking | 0.75 | 1.00 | 0.86 |
| B-Diabetes | 1.00 | 1.00 | 1.00 |
| B-Gender | 0.91 | 0.78 | 0.84 |
| B-Age | 0.58 | 0.56 | 0.57 |
| I-Employment | 0.72 | 0.66 | 0.69 |

An O tag indicates that a token belongs to no entity / chunk. The B-prefix before a tag show that the tag is the beginning of an entity chunk, and an I- prefix before a tag indicates that the tag is inside an entity chunk. The B- tag is used only when a tag is followed by a tag of the same type without O tokens between them. The precision is the ratio of correctly identified named entity identified by our system to the total number of named entities found by the system. The recall is the ratio of correctly identified entity to the total that you should be found. The F1 measure given in the is just the harmonic mean of these two. As seen in Table 6, our model is quite accurate in identifying these entities, which is indicated by quite high scores returned by the model evaluation.

## 6. DISCUSSION

*Practical Impact:* These findings have many implications in the field of healthcare. The clinical evaluation and human evaluation can assist doctors, nurses, and clinical experts in matching symptoms to diagnosis, treatment, and follow-up. The policymakers can use this pipeline to understand the value contained in clinical records for further health system planning. Through this solution, tracking SDoH is also possible that can lead to better clinical outcomes and ultimately leads to the the reduction of health disparities. From the ML perspective, this pipeline is generalizable across different other domains, such as for to study health science echo system, cybersecurity, IOT for patient monitoring, equipment maintenance, and for other life sciences.

*Limitations*: This work is based on the curation of case reports and the biases related to study eligibility criteria, identification and selection of studies can be a limitation. So far, we chose only English as the language, which may have omitted many useful literatures from the corpus. One direction, in this regard is to have rigorous research methods to determines the quality of literature.

For now, we don't have access to any EHR, so we cannot determine the validity of the de-identification component, we have uses fake identifiers to simulate the de-identification process, which suffice the purpose but does not represent the real-time patients' data. In future, we intend to curate more clinical data; getting real-time access to EHRs would be helpful, but we would need to ensure robust de-identification. Since we are already providing a de-identifier to de-identify patients 'personal information through this pipeline, we hope to gain access to such a dataset while adhering to PHIPA guidelines.

## 7. CONCLUSION

In conclusion, this paper presents a ML pipeline for biomedical NER task that consists of a number of nodes stacked together. We use BiLSTM-CNN-CRF model plus BERT-based embeddings to detect biomedical entities. The results show that using contextualized word embedding pre-trained on biomedical corpora significantly improves the results. We evaluated the performance of our approach on five datasets (four benchmark datasets and one own developed case reports dataset) and our approach achieves the best results compared to the baselines. Further, through human evaluation, we conclude that the predicted entities are accurate, informative and biomedical.

In future, we plan to increase the number of layers in this deep neural network for biomedical NER task. We also intend to use a Transformer-based model [26] but first we need to prepare more data for annotation. We strongly encourage the inclusion of medical professionals in the annotation guideline.

## 8. ACKNOWLEDGMENTS

We like to acknowledge the Canadian Institutes of Health Research's Institute of Health Services and Policy Research (CIHR-IHSPR) as part of the Equitable AI and Public Health cohort, and Public Health Ontario.